\documentclass[conference]{IEEEtran}
\IEEEoverridecommandlockouts
\usepackage{cite}
\usepackage{amsmath,amssymb,amsfonts}
\usepackage{algorithmic}
\usepackage{graphicx}
\usepackage{textcomp}
\usepackage{xcolor}
\usepackage{hyperref}
\def\BibTeX{{\rm B\kern-.05em{\sc i\kern-.025em b}\kern-.08em
    T\kern-.1667em\lower.7ex\hbox{E}\kern-.125emX}}

\hypersetup{
    colorlinks=true,
    linkcolor=blue,
    urlcolor=blue,
    citecolor=red,
}
\begin{document}
\title{Robot Conga: A Leader-Follower Walking Approach to Sequential Path Following in Multi-Agent Systems\\
\thanks{All the authors belong to Cyber-Physical Systems, Indian Institute
of Science (IISc), Bengaluru. \{pranavtiwari, soumyodiptan\}@iisc.ac.in 
\\ * denotes equal contribution.}
}

\author{
\IEEEauthorblockN{Pranav Tiwari*, Soumyodipta Nath*}
\IEEEauthorblockN{\href{https://robot-conga.github.io}{https://robot-conga.github.io}}
}

\maketitle

\begin{abstract}
Coordinated path following in multi-agent systems is a key challenge in robotics, with applications in automated logistics, surveillance, and collaborative exploration. Traditional formation control techniques often rely on time-parameterized trajectories and path integrals, which can result in synchronization issues and rigid behavior. In this work, we address the problem of sequential path following, where agents maintain fixed spatial separation along a common trajectory, guided by a leader under centralized control. We introduce Robot Conga, a leader-follower control strategy that updates each agent’s desired state based on the leader’s spatial displacement rather than time, assuming access to a global position reference—an assumption valid in indoor environments equipped with motion capture, vision-based tracking, or UWB localization systems. The algorithm was validated in simulation using both TurtleBot3 and quadruped (Laikago) robots. Results demonstrate accurate trajectory tracking, stable inter-agent spacing, and fast convergence, with all agents aligning within 250 time steps (approx 0.25 seconds) in the quadruped case, and almost instantaneously in the TurtleBot3 implementation. 
\end{abstract}


\section{Introduction}
The multi-agent systems have been a very active area of research since the past few decades. The vast interest in multi-agent distributed or centralized control is vastly driven by it's potential large-scale application in military such as unmanned arial vehicles \cite{9781626}, transportation-based robots \cite{9356608}, satellite formation flying \cite{article_space}, synchronization of coupled oscillations \cite{6018994}, microeconomics \cite{LAMBSON1984282}, power grids \cite{9108633} and so on. 

A key challenge in multi-agent robotics lies in achieving coordinated motion while maintaining specified spatial relationships between agents. Traditional approaches to formation control often rely on time-parameterized trajectories \cite{AILON2012}, which can suffer from synchronization issues, especially when agents experience actuation delays or environmental disturbances. These methods can lead to rigid and brittle behavior, especially in dynamic or uncertain environments.

In this work, we address the problem of sequential path following, where multiple agents are required to move along a shared trajectory while maintaining fixed spatial separations. Unlike rigid formation control, the focus here is on consistent spacing rather than preserving a fixed geometric shape. This problem arises in many real-world scenarios where smooth, orderly traversal through constrained environments is needed.

To address this, we propose Robot Conga, a centralized leader-follower control strategy in which each follower updates its desired state based on the spatial displacement of the leader along the path—rather than time. By decoupling trajectory progression from time, the method naturally avoids desynchronization issues and ensures smoother coordination, even among heterogeneous robots with differing dynamics.

This spatial coordination framework is particularly suited to indoor environments equipped with global localization infrastructure such as motion capture systems, vision-based tracking \cite{9438708, s16122164, pachter2007vision}, inertial measurement units (IMUs) \cite{8606925}, or UWB localization systems \cite{8115911}. Under this assumption, each agent has access to its position in a global frame, enabling precise spatial updates without requiring time synchronization or peer-to-peer communication.

\textbf{Example use cases include:}
\begin{itemize}
    \item Indoor warehouse automation, where delivery bots move along narrow aisles in a train-like manner to fetch or deliver goods.
    \item Mobile inspection in industrial plants, where multiple sensor-equipped robots survey pipelines or factory infrastructure while avoiding congestion.
    \item Guided tours in museums or exhibitions, where a lead robot conducts visitors while follower robots carry assistive technology, supplies, or displays.
    \item Coordinated disinfection in hospitals or airports, where robots follow a prescribed cleaning route with regulated spacing to ensure full coverage.
    \item Search and rescue missions in indoor disaster zones, where a lead robot explores the path and followers carry resources or communication equipment.
\end{itemize}

In such applications, rigid formation control may be unnecessary or even impractical. Instead, ensuring that each robot follows the same trajectory with temporal or spatial offset becomes the central objective.

We validate the proposed Robot Conga algorithm in simulated environments using two distinct robotic platforms: the wheeled TurtleBot3 and a quadruped (Laikago) robot. Results demonstrate accurate path tracking, consistent inter-agent spacing, and rapid convergence. Compared to existing methods based on time-parameterized trajectories and path integrals \cite{AILON2012}, our approach avoids the accumulation of synchronization errors and is better suited to heterogeneous teams and infrastructure-assisted deployment.

\begin{table}[htbp]
\caption{Notation Used in the Paper}
\begin{center}
\begin{tabular}{|c|c|c|c|}
\hline
\textbf{Symbol} & \multicolumn{3}{|c|}{\textbf{Description}} \\
\hline
$x_1(t), x_2(t)$ & \multicolumn{3}{c|}{Position coordinates of the robot at time $t$} \\
\hline
$x_3(t)$ & \multicolumn{3}{c|}{Orientation (heading angle) of the robot at time $t$} \\
\hline
$u(t), \omega(t)$ & \multicolumn{3}{c|}{Linear and angular velocity inputs} \\
\hline
$x_1^*(t), x_2^*(t)$ & \multicolumn{3}{c|}{Desired (reference) position coordinates} \\
\hline
$x_3^*(t)$ & \multicolumn{3}{c|}{Desired (reference) orientation} \\
\hline
$u^*(t), \omega^*(t)$ & \multicolumn{3}{c|}{Reference linear and angular velocity inputs} \\
\hline
$e_1(t), e_2(t), e_3(t)$ & \multicolumn{3}{c|}{Tracking errors in position and orientation} \\
\hline
$\lambda_1, \lambda_2, \lambda_3$ & \multicolumn{3}{c|}{Control gains for feedback law} \\
\hline
$V(e, t)$ & \multicolumn{3}{c|}{Lyapunov function used for stability analysis} \\
\hline
$g(x)$ & \multicolumn{3}{c|}{Parameterized path function} \\
\hline
\multicolumn{4}{l}{$^{\mathrm{a}}$All time-dependent variables are functions of $t$ unless otherwise stated.}
\end{tabular}
\label{tab:notation}
\end{center}
\end{table}

\section{Preliminaries}

In this section, we present the foundational concepts required to understand our coordination strategy. First, we describe the kinematic model and error dynamics of wheeled mobile robots (WMRs). Next, we briefly introduce Lyapunov functions and their relevance in ensuring system stability. 

\subsection{Unicycle Dynamics}

We consider a wheeled mobile robot (WMR) modeled as a unicycle system, which is subject to non-holonomic constraints. The state-space representation is given by:

\begin{equation}
    \begin{aligned}
        \dot{x_1}(t) &= u(t) \cos[x_3(t)] \\
        \dot{x_2}(t) &= u(t) \sin[x_3(t)] \\
        \dot{x_3}(t) &= \omega(t)
    \end{aligned}
\end{equation}

where $(x_1(t), x_2(t))$ denote the robot's position, $x_3(t)$ is the orientation, and $u(t), \omega(t)$ are the linear and angular velocity inputs, respectively.

We define a virtual robot (reference agent) whose desired trajectory and control inputs are given by:

\begin{equation}
    \begin{aligned}
        \dot{x_1}^*(t) &= u^*(t) \cos[x_3^*(t)] \\
        \dot{x_2}^*(t) &= u^*(t) \sin[x_3^*(t)] \\
        \dot{x_3}^*(t) &= \omega^*(t)
    \end{aligned}
\end{equation}

The control objective is to design inputs $u(t)$ and $\omega(t)$ such that the real robot's trajectory converges asymptotically to that of the virtual robot, i.e.,

\[
\{x_1(t), x_2(t), x_3(t)\} \rightarrow \{x_1^*(t), x_2^*(t), x_3^*(t)\} \quad \text{as} \quad t \rightarrow \infty
\]

This convergence ensures that each agent accurately tracks its assigned trajectory segment as determined by the centralized coordination scheme.

\subsection{Error Dynamics and Control Law}

To ensure the real robot tracks the desired trajectory, we define the tracking error between the actual and reference states as:

\begin{equation}
    \mathbf{e}(t) = 
    \begin{bmatrix}
        e_1(t) \\
        e_2(t) \\
        e_3(t)
    \end{bmatrix}
    =
    \begin{bmatrix}
        x_1(t) - x_1^*(t) \\
        x_2(t) - x_2^*(t) \\
        x_3(t) - x_3^*(t)
    \end{bmatrix}
\end{equation}

Taking the time derivative of the error vector, we obtain the error dynamics:

\begin{equation}
    \begin{bmatrix}
        \dot{e}_1(t) \\
        \dot{e}_2(t) \\
        \dot{e}_3(t)
    \end{bmatrix}
    =
    \begin{bmatrix}
        u(t) \cos[x_3(t)] - u^*(t) \cos[x_3^*(t)] \\
        u(t) \sin[x_3(t)] - u^*(t) \sin[x_3^*(t)] \\
        \omega(t) - \omega^*(t)
    \end{bmatrix}
\end{equation}

\subsection{Lyapunov Functions and Asymptotic Stability}

Lyapunov theory provides a powerful tool to assess the stability of dynamical systems without requiring the explicit solution of differential equations. The fundamental idea is to construct a scalar function, called a \textit{Lyapunov function}, that acts like an energy measure of the system.

\subsubsection*{Definition}

Consider a nonlinear system of the form:
\begin{equation}
    \dot{x} = f(x), \quad x \in \mathbb{R}^n, \quad f(0) = 0
\end{equation}
where the origin is an equilibrium point. A continuously differentiable function $V: \mathbb{R}^n \to \mathbb{R}$ is said to be a Lyapunov function if it satisfies the following properties in a neighborhood $\mathcal{D}$ of the origin:

\begin{itemize}
    \item $V(0) = 0$ and $V(x) > 0$ for all $x \in \mathcal{D} \setminus \{0\}$ (positive definite),
    \item $\dot{V}(x) = \frac{dV}{dt} = \nabla V(x)^T f(x) \leq 0$ for all $x \in \mathcal{D}$ (negative semi-definite).
\end{itemize}

\subsubsection*{Stability Implications}

\begin{itemize}
    \item If such a Lyapunov function exists, the equilibrium at the origin is \textit{Lyapunov stable}.
    \item If $\dot{V}(x) < 0$ for all $x \neq 0$, the equilibrium is \textit{asymptotically stable}.
    \item If $V(x) \to \infty$ as $\|x\| \to \infty$ (i.e., $V$ is radially unbounded), and $\dot{V}(x) < 0$ globally, then the system is \textit{globally asymptotically stable}.
\end{itemize}

\subsubsection*{Why Lyapunov Functions Matter}

Lyapunov-based methods are widely used in robotics and control because they allow for the design and verification of controllers that guarantee stability, even in the presence of nonlinearities. Instead of solving the system’s equations of motion, we analyze the behavior of a carefully constructed scalar function whose decrease over time implies convergence to a desired state.

This framework will be used in later sections to verify the stability of our proposed control strategy.

\section{Stability of the System}

To analyze the stability of the closed-loop system, we adopt a Lyapunov-based approach similar to the one proposed in \cite{AILON2007499}. 

\begin{equation}
    V(e) = \frac{1}{2} e^T
    \begin{bmatrix}
        \delta & 0 & 0 \\
        0 & \delta_1 & 1 \\
        0 & 1 & 1
    \end{bmatrix}
    e
\end{equation}

where \( \delta > 0 \), \( \delta_1 > 0 \) are constants, and \( e = [e_1, e_2, e_3]^T \) is the error vector defined earlier.

\subsection{Time Derivative of the Lyapunov Function}

The time derivative of \( V(e) \) along the trajectories of the system is:

\begin{equation}
\begin{aligned}
    \dot{V}(e, t) &= \delta e_1 \dot{e}_1 + \delta_1 e_2 \dot{e}_2 + \dot{e}_2 e_3 + e_2 \dot{e}_3 + e_3 \dot{e}_3 \\
    &= -\delta \lambda_3 e_1^2 - \delta_1 \sigma e_1 e_2 - \sigma e_1 e_3  \\
    &\quad + \delta_1 \psi e_2 \sin(e_3) + \psi e_3 \sin(e_3) \\
    &\quad - a e_2^2 - b e_3^2 - a e_2 e_3 - b e_2 e_3
\end{aligned}
\end{equation}

where
\[
\sigma(e_3, t) = \lambda_3 \tan(e_3 + x_3^*(t)), \quad \psi(e_3, t) = \frac{u^*(t)}{\cos(e_3 + x_3^*(t))}.
\]

\subsection{Control Law}

We employ control approach similar to the one proposed in \cite{AILON2007499} :

\begin{equation}
    \begin{aligned}
        u(t) &= \frac{u^*(t) \cos[x_3^*(t)] - \lambda_3 e_1}{\cos(x_3^*(t) + e_3 )}, \\
        \omega(t) &= \omega^*(t)  - \lambda_2 e_3 - \lambda_1 e_2,
    \end{aligned}
\end{equation}

where \( \lambda_1, \lambda_2, \lambda_3 > 0 \) are control gains.

\subsection*{Resulting Error Dynamics}

Substituting the control law into the system yields the following closed-loop error dynamics:

\begin{equation}
\begin{bmatrix}
    \dot{e}_1 \\
    \dot{e}_2 \\
    \dot{e}_3
\end{bmatrix}
=
\begin{bmatrix}
    -\lambda_3 e_1 \\
    -\lambda_3 e_1 \tan(x_3^*) + \dfrac{u^* \sin(e_3)}{\cos(e_3 + x_3^*)} \\
    -\lambda_1 e_2 - \lambda_2 e_3
\end{bmatrix}
\label{eq:suberrordynamics}
\end{equation}

\subsection*{Exponential Stability}

As shown in \cite{AILON2007499}, for any \( \lambda_3 > 0 \), one can choose positive constants \( \delta, \delta_1, \lambda_1, \lambda_2 \) such that the Lyapunov function \( V(e) \) is positive definite and its derivative \( \dot{V}(e) \) is negative definite in a neighborhood \( D \subset \mathbb{R}^3 \), i.e.,

\[
\dot{V}(e) < 0 \quad \forall e \in D.
\]

Moreover, there exist positive constants \( \alpha, \beta, \rho \) such that:

\begin{equation}
    \alpha \|e\|^2 \leq V(e) \leq \beta \|e\|^2, \quad \dot{V}(e) \leq -\rho \|e\|^2,
\end{equation}

implying that the origin is an exponentially stable equilibrium point of the system in \( D \).

Thus, the proposed control law guarantees that the real robot's trajectory converges to the virtual reference trajectory, ensuring asymptotic stability of the system.

\section{Trajectory Propagation and Real-Time Adaptation in Leader-Follower Systems}

In our proposed framework, the leader's movement along the path determines the reference positions and velocities for all follower robots. By using arc-length-based updates instead of time-parameterized trajectories, we ensure that inter-bot distances remain consistent and synchronization issues are avoided. The approach combines spatial propagation of reference states and real-time path updates through user interaction. This dual-layered strategy ensures scalability, adaptability, and smooth navigation in dynamically changing environments.

\subsection{Spatial Propagation of Reference States}

\begin{figure}[htbp]
    \centering
    \includegraphics[width=0.48\textwidth]{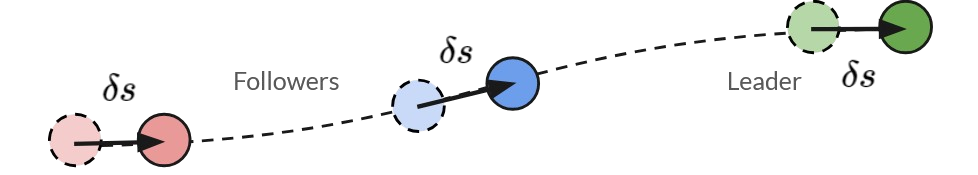}
    \caption{Displacement-based State Update}
    \label{fig:botmovements}
\end{figure}

Whenever the leader robot advances a small displacement \( \left( \delta s \right) \) along the path, the desired states \( \left(x_{i1}^*(t), x_{i2}^*(t), x_{i3}^*(t)\right) \) of each follower robot are updated proportionally to maintain equal spacing  along the curve, where i denotes the index of the robot in the formation (with $i=1$ being the leader and $i>1$ referring to followers) as showncased in Figure \ref{fig:botmovements}.

Assuming that the desired linear velocity of all bots along the trajectory is equal and is controlled in real-time via joystick input.

\[
u^*_{i} = {\mathbb{V}_{cmd}}
\]

We discretize the trajectory propagation for real-time implementation by updating the desired states of each follower at discrete time steps. Let \(\mathbf{k}\) denote the discrete time step, and \(\delta t\) be the time interval between updates. The updated desired states at each time step are given by:

\begin{equation}
\begin{aligned}
    \dot{x}_{i1}^* = \dot{u}_{i}^* cos(x_{i3}^*)  
    \Rightarrow {x}_{i1}^*[k] &= {x}_{i1}^*[k-1]  + \dot{u}_{i} \times cos(x_{i3}^*[k-1]) \times\delta t \\
    x_{i2}^*[k] &= g(x_{i1}^*[k]) \\
    x_{i3}^*[k] &= \left. \tan^{-1}\left( \frac{dg(x_1)}{dx_1} \right) \right|_{x_1 = x_{i1}^*[k]}
\end{aligned}
\end{equation}

The desired angular velocities are determined on the basis of the instantaneous radius of curvature (IROC) of the path at each bot's location:

\begin{equation}
\omega^*_{i} = \frac{u^*_{i}}{\text{IROC}} = \left| \frac{{\mathbb{V}_{cmd}} \cdot \frac{d^2 g(x_1)}{dx_1^2}}{\left[1 + \left(\frac{dg(x_1)}{dx_1}\right)^2 \right]^{3/2}} \right|_{x_1 = x_{i1}^*[k]}
\end{equation}

\subsection*{Initial Configuration}

\begin{figure}[htbp]
    \centering
    \includegraphics[width=0.48\textwidth]{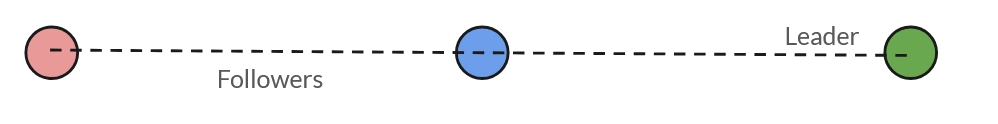}
    \caption{Leader-Follower Initialization}
    \label{fig:botinit}
\end{figure}

At \( \mathbf{k} = 0 \), the follower robots are initialized at fixed arc-length intervals behind the leader along the curve. This initialization ensures uniform propagation as the leader moves as shown in Figure \ref{fig:botinit}.

\subsection{Dynamic Path Adaptation Using Joystick Steering}

To further enhance responsiveness, we incorporate joystick or keyboard-based steering to dynamically adjust the leader robot’s heading. This allows the formation to adapt dynamically to new goals or environmental constraints. 

\begin{figure}[htbp]
    \centering
    \includegraphics[width=0.48\textwidth]{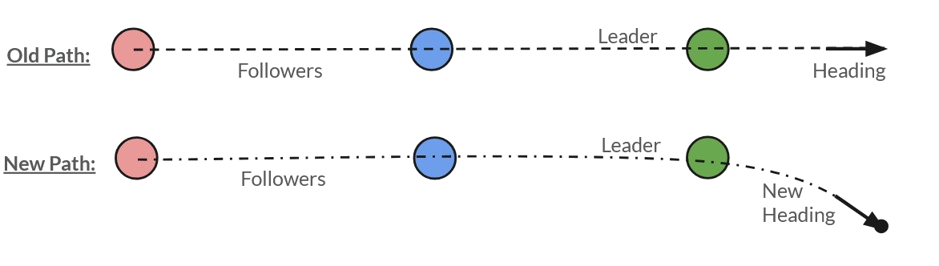}
    \caption{Key-Controlled Dynamic Trajectory}
    \label{fig:keycontoltraj}
\end{figure}

Whenever the leader’s heading is modified, the original path is no longer valid beyond the current point. Therefore, we recalculate a new reference trajectory \(g^*(x)\) that smoothly transitions from the robot’s current position and heading to the intended direction (Figure~\ref{fig:keycontoltraj}).

To generate this updated trajectory, we interpolate a set of virtual waypoints based on the current pose and the desired steering input. The result is a spatial path that serves as a new reference for both the leader and the follower robots

To do so we interpolate the bot locations 

Several interpolation methods can be used to construct the smooth trajectory from the updated waypoints:

\begin{itemize}
\item \textbf{Barycentric Interpolation} \cite{doi:10.1137/S0036144502417715}: While computationally efficient, this method is highly sensitive to changes in orientation and can result in undesirable global distortions or sharp turns, especially under dynamic user inputs.

\item \textbf{B-Spline Interpolation} \cite{chaudhuri2021b}: This piecewise polynomial method offers local control, ensuring smooth transitions in curvature. It performs robustly under real-time heading changes, producing natural and collision-free trajectories.

\end{itemize}

\begin{figure}[htbp]
\centering
\includegraphics[width=0.48\textwidth]{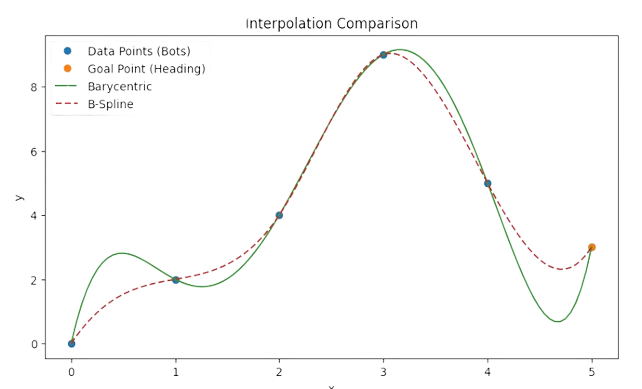}
\caption{Interpolation Method Comparison}
\label{fig:interpolationcomparison}
\end{figure}

Based on comparative performance (Fig~\ref{fig:interpolationcomparison}), B-spline interpolation was selected as the default approach for generating dynamically updated reference trajectories in real time. Its smooth curvature and local adaptability make it ideal for human-in-the-loop navigation scenarios.

\subsection{Trajectory Correction on Steep Ascent}
Since the angular speed ($\omega$) depended on the slope of the trajectory, the measurement would become unstable as the trajectory aligned with the Y-axis—that is, when the slope approached $\infty$. To mitigate this issue, a mechanism was implemented to allow a frame change whenever the heading angle began aligning with the Y-axis. This caused the local frame to rotate, updating the coordinates in the local frame and ensuring that the trajectory would never become parallel to the Y-axis at any point.

\section{Results}

To evaluate the performance, robustness, and versatility of the proposed \textit{Robot Conga} algorithm, we conducted experiments across a diverse set of robotic platforms: wheeled robots (TurtleBot3), legged robots (Laikago Quadrupeds), and heterogeneous combinations of both. The architecture abstracts away low-level actuation differences and requires tuning only high-level parameters \( \lambda_1, \lambda_2, \lambda_3 \) for stable convergence and consistent formation behavior.

Table~\ref{tab:lambda_values} summarizes the values of these control parameters for each tested configuration.

\begin{table}[htbp]
\caption{Control Parameters Used for Different Robot Types}
\begin{center}
\begin{tabular}{|c|c|c|c|}
\hline
\textbf{Robot Type} & \(\boldsymbol{\lambda_1}\) & \(\boldsymbol{\lambda_2}\) & \(\boldsymbol{\lambda_3}\) \\
\hline
TurtleBot3 & 4.5 & 7.5 & 2.5 \\
\hline
Laikago Quadruped & 4.5 & 1.5 & 2.5 \\
\hline
Heterogeneous (Mixed) & 5.0 & 1.0 & 1.5 \\
\hline
\end{tabular}
\label{tab:lambda_values}
\end{center}
\end{table}

\subsection{TurtleBot3 Implementation}

The algorithm was implemented in ROS2 \cite{Ros2} using multiple TurtleBot3 robots\footnote{\url{https://wiki.ros.org/turtlebot3}} within the Gazebo simulation environment \cite{gazebo}, where each robot was spawned under a unique namespace. A keyboard-based interface controlled the heading and speed of the leader in real time, while follower bots adapted using arc-length-based spatial propagation.

The system's performance was highly sensitive to the tuning of high-level control parameters \( \lambda_1, \lambda_2, \lambda_3 \), commanded linear velocity \( \mathbb{V}_{cmd} \), and heading adjustment per keystroke. Each keypress served as a discrete perturbation to the leader’s trajectory, introducing a transient increase in error across the formation. However, due to the convergence properties of the control law, both positional and angular tracking errors consistently diminished over time.

\begin{figure}[htbp]
    \centering
    \includegraphics[width=0.48\textwidth]{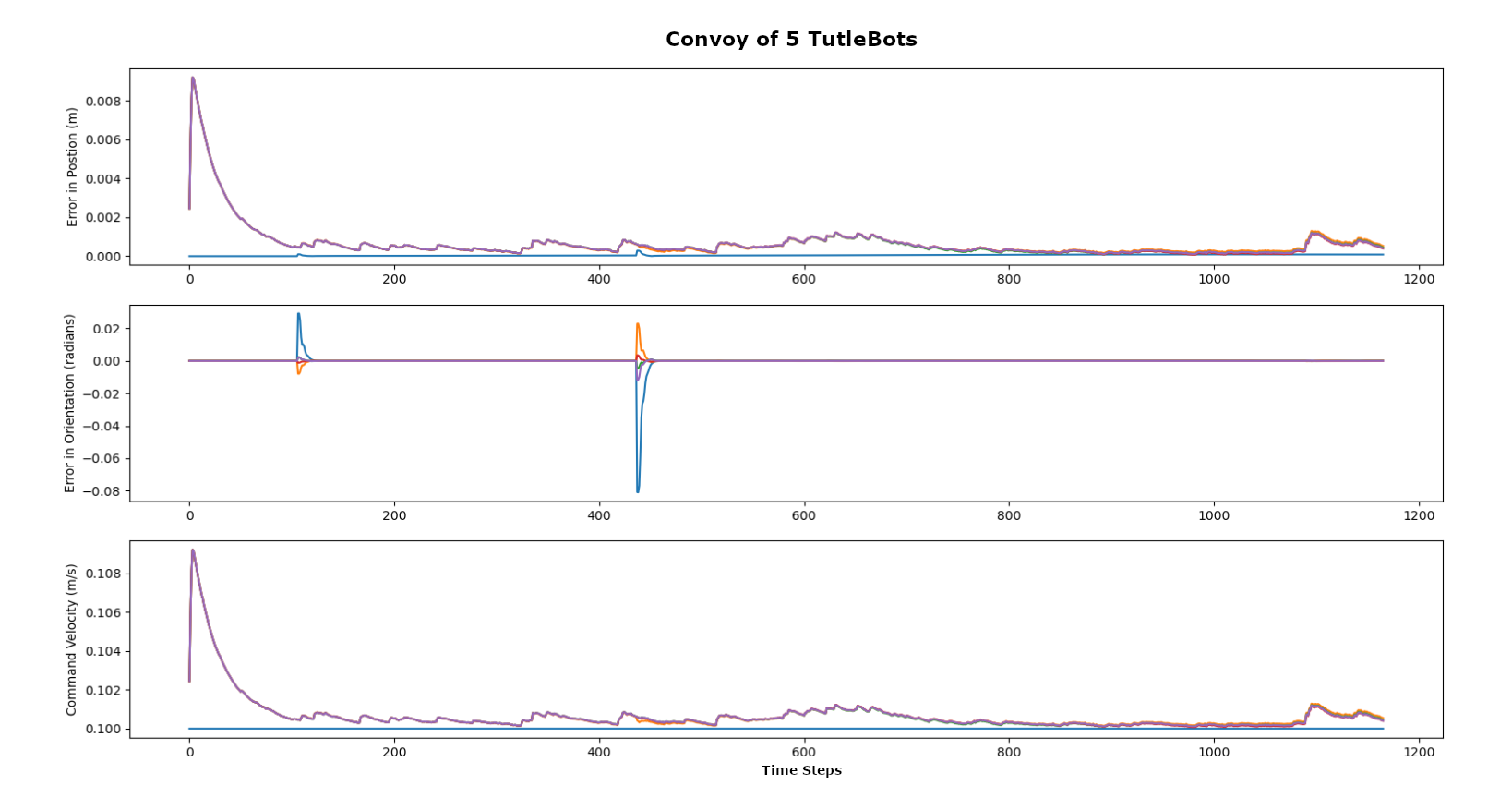}
    \caption{Tracking Performance on TurtleBot3: Positional and angular errors converge despite dynamic path changes. [Multiple spikes represent individual direction and velocity commands]}
    \label{fig:results_tb3}
\end{figure}

Notably:
\begin{itemize}
    \item \textbf{Linear velocity convergence}: Follower bots reached the commanded linear velocity almost instantaneously.
    \item \textbf{Angular convergence}: Angular deviations were corrected within a few milliseconds post-perturbation.
    \item \textbf{Asymptotic stability}: Tracking errors reduced smoothly and converged nearly to zero as shown in Figure~\ref{fig:results_tb3}.
\end{itemize}

Figure~\ref{fig:results_tb3} demonstrates this stability, even under frequent trajectory modifications introduced by user input. A video demonstration is available online\footnote{TurtleBot3 experiment video: \url{https://www.youtube.com/watch?v=b6I9cNeFR_4}}. \footnote{Project Github Codebase: \url{https://github.com/Tiwari-Pranav/Robot-conga-turtlebot3-ros-gazebo}}

\subsection{Laikago Quadruped Implementation}

\begin{figure}[htbp]
    \centering
    \includegraphics[width=0.48\textwidth]{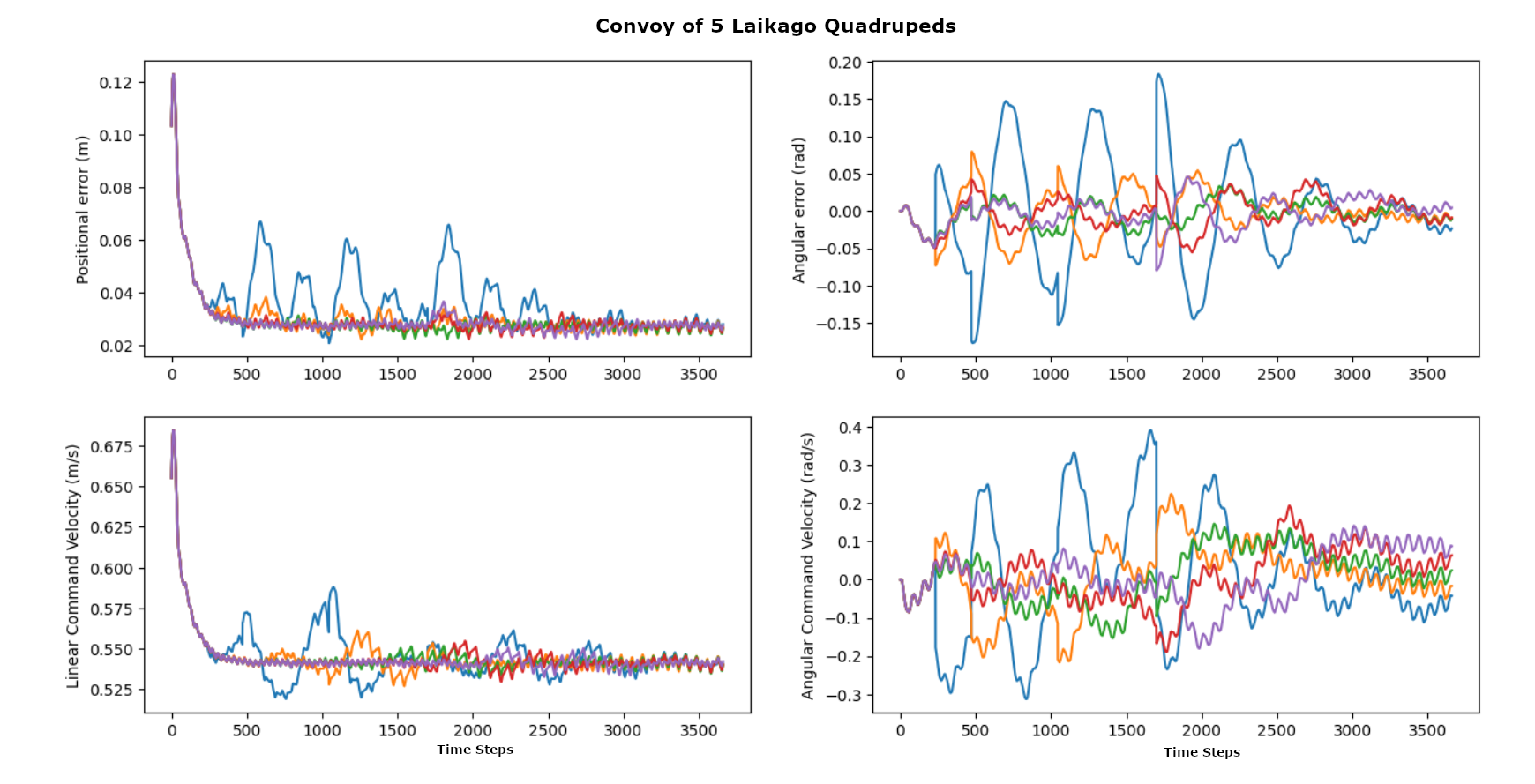}
    \caption{Tracking Performance on Laikago Quadrupeds: Rapid damping of trajectory error following steering input. [Multiple spikes represent individual direction and velocity commands]}
    \label{fig:results_quadruped}
\end{figure}

To validate the generalizability of our approach, we implemented the algorithm on Laikago quadruped robots in the PyBullet simulation environment \cite{coumans2016pybullet}. These robots use a low-level Model Predictive Control (MPC) system\footnote{https://github.com/Farama-Foundation/a2perf-quadruped-locomotion} , while our high-level framework provided reference updates based on the leader's motion.

Despite the platform's drastically different dynamics, the system demonstrated rapid error convergence. As shown in Figure~\ref{fig:results_quadruped}, each joystick-induced heading change caused a spike in error, which stabilized within 250 simulation steps (approx 0.25 seconds). This response underscores the algorithm’s robustness under sudden direction shifts and its compatibility with legged locomotion systems. A demonstration video is also available\footnote{Laikago experiment video: \url{https://www.youtube.com/watch?v=foqDSqTVpeE}}. \footnote{Project Github Codebase: \url{https://github.com/Tiwari-Pranav/Robot-conga-quadruped-pybullet}}

\subsection{Heterogeneous Multi-Robot Implementation}

\begin{figure}[!htbp]
    \centering
    \includegraphics[width=0.48\textwidth]{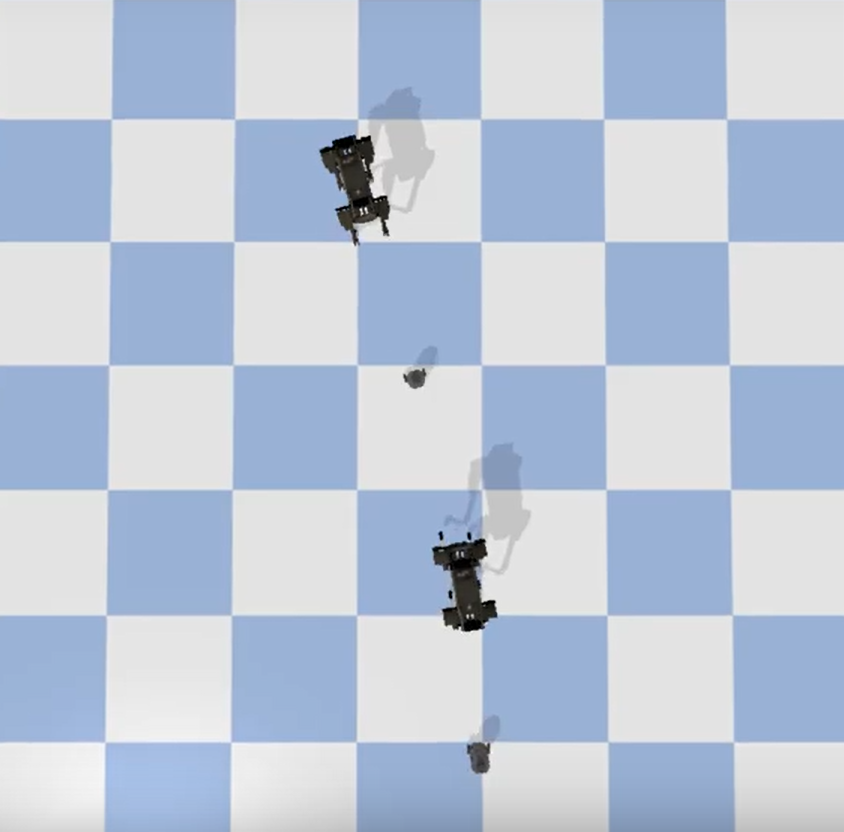}
    \caption{TurtleBot3 and Laikago robots exhibit stable convergence following dynamic path updates.}
    \label{fig:results_heterogeneous_IMG}
\end{figure}

The final experiment evaluated the algorithm’s performance in a mixed-robot setting combining TurtleBot3 (wheeled) and Laikago (legged) platforms (Figure~\ref{fig:results_heterogeneous_IMG}). Despite their differing locomotion and control architectures, both platforms successfully maintained consistent tracking behavior under the same high-level propagation framework.

\begin{figure}[!htbp]
    \centering
    \includegraphics[width=0.48\textwidth]{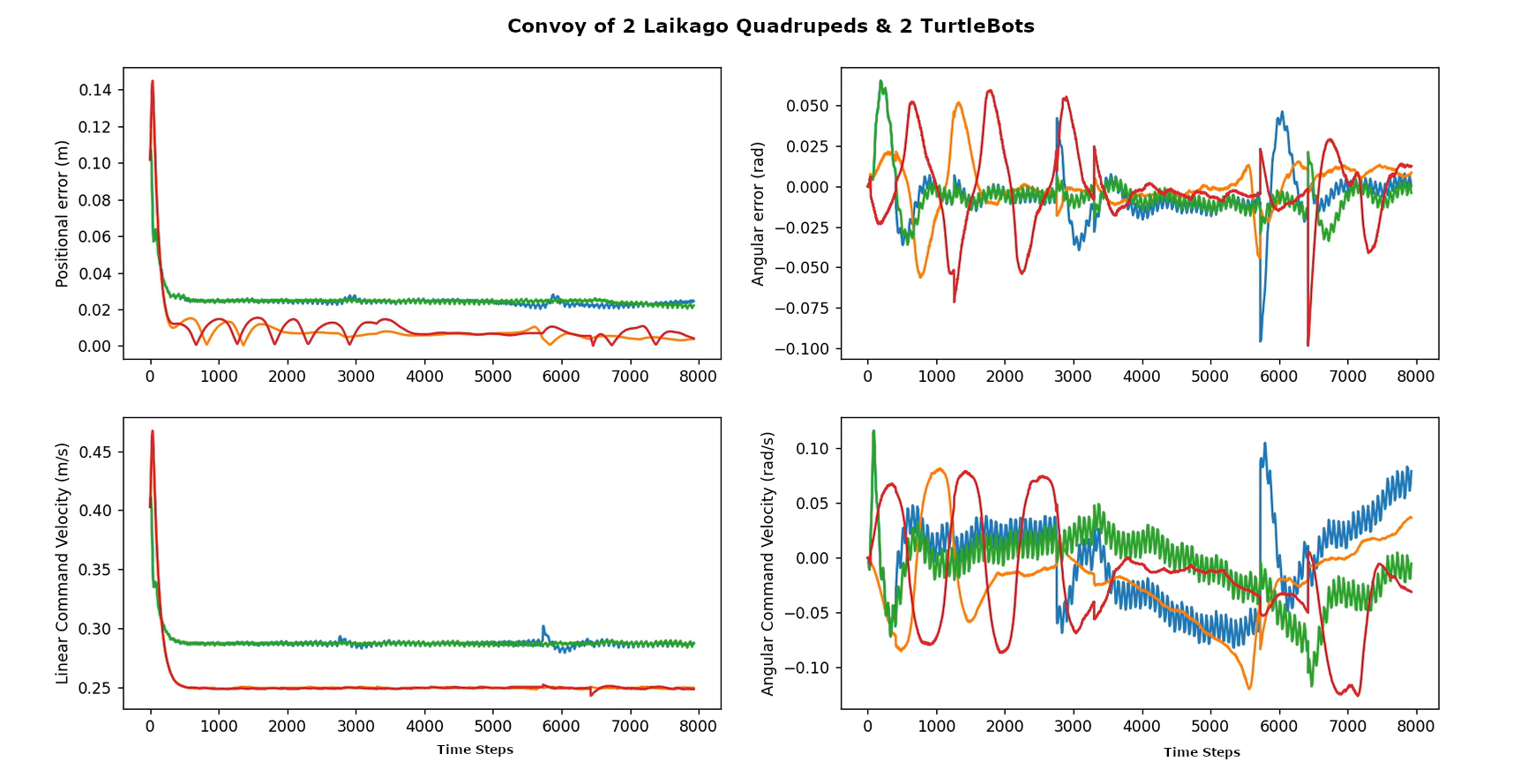}
    \caption{Tracking Error in Mixed-Platform Setup: Both TurtleBot3 and Laikago robots exhibit stable convergence following dynamic path updates. [Multiple spikes represent individual direction and velocity commands]}
    \label{fig:results_heterogeneous}
\end{figure}

Figure~\ref{fig:results_heterogeneous} shows the evolution of tracking error over time for both robots. Error peaks correspond to steering commands, while rapid convergence back to baseline demonstrates the stability and coordination effectiveness of the proposed method.

This result highlights the platform-agnostic nature of the control architecture: although the physical actuation mechanisms differ, both robots interpret the propagated reference consistently. A video of this experiment is available\footnote{Heterogeneous robot experiment video: \url{https://www.youtube.com/watch?v=A-Nygq5zwCc}}.

\section{Conclusion}

In this work, we introduced a spatial propagation-based control framework for trajectory tracking in multi-robot systems, termed the \textit{Robot Conga} algorithm. By decoupling trajectory propagation from time and instead anchoring it to the leader’s displacement, we maintained fixed inter-bot spacing and achieved synchronization across heterogeneous platforms. The proposed method was demonstrated to be adaptable, robust, and scalable across both wheeled and legged robots, with minimal tuning required per platform.

Our results confirmed that the system achieves asymptotic stability even under frequent heading changes and external disturbances such as joystick inputs. Through careful selection of interpolation methods—specifically B-splines—we ensured smooth and dynamic trajectory generation in real time.

The algorithm's abstraction over low-level control layers and its compatibility with both simulated and real-world robotic systems make it a promising candidate for formation control tasks in complex environments.

\section{Future Work}

Several directions remain open for extending the current work. One potential avenue is adapting the proposed algorithm for deployment in GPS-denied environments, using onboard perception and SLAM-based localization to estimate the leader's trajectory and propagate it throughout the formation.

Another promising direction involves incorporating advanced control strategies to address communication and sensing delays. Integrating Observer-Based Feedback Protocols, as discussed in \cite{10993485}, could enhance robustness in networked robotic systems where latency is a critical factor.

Further exploration into Consensus Control frameworks may help achieve prescribed performance guarantees in leader-follower multi-agent systems, as outlined in \cite{9029509}.

Lastly, to improve reliability in real-world deployments, it is valuable to investigate resilient formation control under partial system failures, such as leader dropout. This aligns with ongoing work in leader-failure resilience, including studies such as \cite{Murakami31122025}.

\bibliographystyle{ieeetr}
\bibliography{citations}

\end{document}